\documentclass{article}

\usepackage{microtype}
\usepackage{graphicx}
\usepackage{subcaption}
\usepackage{booktabs} 

\usepackage[ruled,algo2e]{algorithm2e}
\usepackage{xcolor}
\usepackage{float}
\usepackage{amssymb}
\usepackage{amsthm}
\usepackage{scalerel}
\usepackage{stackengine}
\usepackage[nointegrals]{wasysym}
\usepackage{tablefootnote}
\usepackage{wrapfig}
\usepackage{xfrac}

\usepackage{tabularx}
\usepackage{mathtools}
\usepackage{bbm}

\usepackage{anyfontsize}

\usepackage{hyperref}

\usepackage[accepted]{icml2021}

\icmltitlerunning{Augmented World Models Facilitate Zero-Shot Dynamics Generalization From a Single Offline Environment}

\begin{document}

\twocolumn[
\icmltitle{Augmented World Models Facilitate Zero-Shot Dynamics Generalization \\ From a Single Offline Environment}

%



\icmlsetsymbol{equal}{*}

\begin{icmlauthorlist}
\icmlauthor{Philip J. Ball}{equal,ox}
\icmlauthor{Cong Lu}{equal,ox}
\icmlauthor{Jack Parker-Holder}{ox}
\icmlauthor{Stephen Roberts}{ox}
\end{icmlauthorlist}


\icmlaffiliation{ox}{University of Oxford, United Kingdom}

\icmlcorrespondingauthor{Philip Ball}{ball@robots.ox.ac.uk}
\icmlcorrespondingauthor{Cong Lu}{cong.lu@stats.ox.ac.uk}

\icmlkeywords{Machine Learning, ICML, Reinforcement Learning, Offline reinforcement learning, model based reinforcement learning, zero-shot generalization, augmentations, robustness}

\vskip 0.3in
]



\printAffiliationsAndNotice{\icmlEqualContribution} 

\begin{abstract}
Reinforcement learning from large-scale offline datasets provides us with the ability to learn policies without potentially unsafe or impractical exploration. Significant progress has been made in the past few years in dealing with the challenge of correcting for differing behavior between the data collection and learned policies. However, little attention has been paid to potentially changing dynamics when transferring a policy to the online setting, where performance can be up to 90\% reduced for existing methods. In this paper we address this problem with \emph{Augmented World Models} (AugWM). We augment a learned dynamics model with simple transformations that seek to capture potential changes in physical properties of the robot, leading to more robust policies. We not only train our policy in this new setting, but also provide it with the sampled augmentation as a context, allowing it to adapt to changes in the environment. At test time we \emph{learn the context} in a self-supervised fashion by approximating the augmentation which corresponds to the new environment. We rigorously evaluate our approach on over $100$ different changed dynamics settings, and show that this simple approach can significantly improve the zero-shot generalization of a recent state-of-the-art baseline, often achieving successful policies where the baseline fails.
\end{abstract}

\section{Introduction}
\label{sec:intro}


Offline reinforcement learning (RL) describes the problem setting where RL agents learn policies solely from previously collected experience without further interaction with the environment \cite{bcq, offlinerl_survey}. This could have tremendous implications for real world problems \cite{realworldrl}, with the potential to leverage rich datasets of past experience where exploration is either not feasible (e.g. a Mars Rover) or unsafe (e.g. in medical settings). As such, interest in offline RL has surged in recent times. 

This work focuses on model-based offline RL, which has achieved state-of-the-art performance through the use of uncertainty penalized updates \cite{mopo, morel}. However, existing work only addresses the issue of transferring from different behavior policies in the \emph{same environment}, ignoring any possibility of distribution shift. Consider the case where it is expensive to collect data, and we have access to a single dataset from a robot. Using existing methods we would be unable to make any changes that impact the dynamics, such as using a newer model of the robot or deploying it in a different room.

A related setting is the Sim2Real problem which considers transferring an agent from a simulated environment to the real world. A popular recent approach is domain randomization \cite{domain_randomization, james2017transferring}, the process of randomizing non-essential regions of the observation space to make agents robust to `observational overfitting' \cite{Song2020Observational}. Indeed, methods seeking to generalize to novel dynamics have also shown promise \cite{sim2realdynamics}, by randomizing physical properties such as the mass of the agent. A significant limitation of these approaches is the requirement for a simulator, which may not be available. 

\begin{figure*}[ht]
\centering

  ~
  \centering\begin{subfigure}{0.22\textwidth}
    \centering\includegraphics[height=1.8cm]{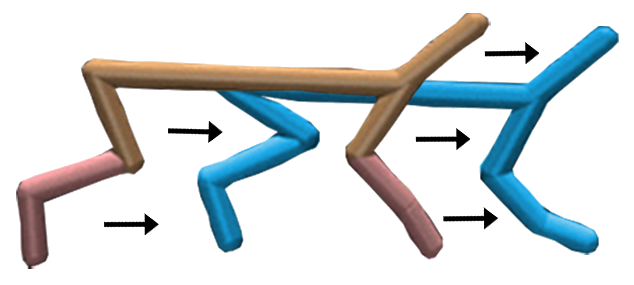}
    \caption{\small{World Model: $\widehat{P}$}}
    \label{fig:dynm1}
  \end{subfigure}
  ~
  \hspace*{1cm}
  \centering\begin{subfigure}{0.3\textwidth}
    \centering\includegraphics[height=1.8cm]{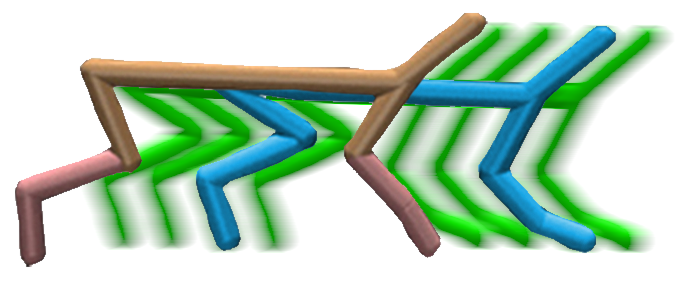}
    \caption{\small{Augmented World Model: $\widehat{P}_z$}}
    \label{fig:dynm2}
  \end{subfigure}
  ~
  \hspace*{0.7cm}
  \centering\begin{subfigure}{0.22 \textwidth}
    \centering\includegraphics[height=1.8cm]{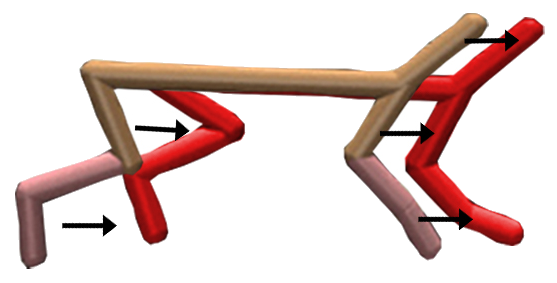}
    \caption{Test: $P^\star$}
    \label{fig:dynm3}
  \end{subfigure}
  \vspace{-3mm}
  \caption{\small{An illustration of our approach, each figure shows a transition $P(s,a)\rightarrow s'$. In a) we show the World Model dynamics {\footnotesize $\widehat{P}$}, trained from $\mathcal{D}_\text{env}$, a single offline dataset. In b) we show the Augmented World Model, blue represents {\footnotesize $\widehat{P}$}, while each green agent illustrates an instantiation of augmented dynamics, which is sampled at each timestep: {\footnotesize $\widehat{P}_z, z\sim \mathcal{Z}$}. The goal is to approximate c) where we show an unseen test environment, with transition dynamics $P^\star$.}}
  \label{fig:dynamics_diff}
  \vspace{-2mm}
\end{figure*}

In this work we take inspiration from Sim2Real to generalize solely from an offline dataset, in a learned simulator or \emph{World Model} (WM). We therefore describe our problem setting as follows: an agent must learn to generalize to unseen test-time dynamics whilst having access to offline data from \emph{only a single environment}; we call this ``dynamics generalization from a single offline environment". 

In this paper we concentrate on the zero-shot performance of our agents to unseen dynamics, as it may not be practical (nor safe) to perform multiple rollouts at test time. To tackle this problem, we propose a novel form of data augmentation: rather than augment observations, we focus on \emph{augmenting the dynamics}. We first learn a world model of the environment, and then augment the transition function at policy training time, making the agent train under different imagined dynamics. In addition, our agent is given access to the augmentation itself as part of the observations, allowing it to consider the context of modified dynamics.

At test time we propose a simple, yet surprisingly effective, self-supervised approach to learning an agent's augmentation context. We learn a linear dynamics model which is then used to approximate the dynamics augmentation induced by the modified environment. This context is then given to the agent, allowing it to adapt on the fly to the new dynamics within a single episode (i.e. zero-shot).
We show that our approach is capable of training agents that can vastly outperform existing Offline RL methods on the ``dynamics generalization from a single offline environment" problem. We also note that this approach does \emph{not} require access to environment rewards at test time. This facilitates application to Sim2Real problems whereby test time rewards may not be available.

Our contributions are twofold: 1) As far as we are aware, we are the first to propose \textbf{dynamics augmentation for model based RL}, allowing us to generalize to changing dynamics despite only training on a single setting. We do this without access to any environment parameters or prior knowledge. 2) We propose a simple \textbf{self-supervised context adaptation} reward-free algorithm, which allows our policy to use information from interactions in the environment to vary its behavior in a single episode, increasing zero-shot performance. We believe both of these approaches are not only novel, but offer significant improvement v.s.\ state-of-the-art methods, improving generalization and providing a promising approach for using offline RL in the real world.

\section{Related Work}

In this work we focus on Model Based RL (MBRL). A key challenge in MBRL is that an inaccurate model can be exploited by the policy, leading to behaviors that fail to transfer to the real environment. As such, a swathe of recent works have made use of model ensembles to improve robustness \cite{METRPO, pets, mbmpo, mbpo, rp1}. With increased accuracy, MBRL has recently been shown to be competitive with model free methods in continuous control \cite{worldmodels, pets, mbpo} and games \cite{muzero, simple}. We make use of an ensemble of probabilistic dynamics models, first introduced in \citet{deepprobensembles} and subsequently used in \citet{pets}.

In this paper we focus on Model-Based \emph{offline} RL, where MOPO \cite{mopo} and MOReL \cite{morel} have recently demonstrated the effectiveness of learned dynamics models, using model uncertainty to constrain policy optimization. We build upon this approach for zero-shot dynamics generalization from offline data. There have also been successes in off policy methods for offline RL \cite{brac, bcq, cql, rudner2021pathologies} and context based approaches \cite{ajay2021opal}, although these works only consider tasks within the support of the offline dataset. Finally, MBOP \cite{argenson2021modelbased} addresses the problem of goal-conditioned zero-shot transfer from offline datasets. However, their goal-conditioning relies on unchanged dynamics in the test environment.

In online RL, recent work has achieved strong dynamics generalization with a learned model \cite{tmcl}. However, this required training under varied dynamics, assigning different experiences to models. In addition, this work used MPC whereas we train a policy inside the model, which is significantly faster at deployment time. Also related are \citet{clavera2018learning, nagabandi2018deep}, where the model is trained to quickly to adapt to new dynamics $P(s'|s,a)$, however both these works place more emphasis on model-adaption rather than zero-shot policy performance. Furthermore, access to an underlying task distribution is required, something we do not have in our offline setting. Also similar to our work is the recently proposed Policy Adaptation during Deployment (PAD, \cite{hansen2021selfsupervised}) approach. Our approach differs in that we learn a context, whereas PAD uses a auxiliary objective to adapt its features. In addition, PAD considers the \emph{online model free} setting, while our method is \emph{offline and model based}. 

Sim2Real is the setting where an agent trained in a simulator must transfer to the real world. A common approach to solve this problem is through domain randomization \cite{domain_randomization, james2017transferring}, whereby parameters in the simulator are varied during training. This has shown to be effective for dynamics generalization \cite{dexterity, antonova2017reinforcement,sim2realdynamics, dynamics_gen_sim2real, zhou2018environment, rubics_cube}, but requires access to a simulator which we do not have. Another form of domain randomization, data augmentation, has proved to be effective for training RL policies \cite{rad, curl, drq, ucb_drac}, resulting in improved efficiency and generalization. So far, these works have focused on online model free methods, and used data augmentation on the \emph{state} space, reducing \emph{observational} overfitting \cite{Song2020Observational}. In contrast, we focus on offline MBRL and instead augment the \emph{dynamics}.

We also note clear links to contextual MDPs \cite{hallak2015contextual, pmlr-v83-modi18a} and hidden parameter MDPs (HiP-MDP) \cite{hipmdp2016doshivelez,Killian_Konidaris_Doshi-Velez_2017, zhang2021learning} settings, whereby our self-supervised dynamics embedding can be considered as a context/hidden parameter. However in these settings the embedding is chosen at the beginning of each episode and is fixed throughout, whereas our embedding varies per timestep.

We are not the first to propose data augmentation in the MBRL setting, \citet{pitis2020counterfactual} proposed Counterfactual data augmentation for improving performance in the context of locally factored tasks. Approaches to ensuring adversarial robustness can include data augmentations that assist with out-of-domain generalization, as opposed to observational overfitting. In \citet{volpi2018unseen} this is done without a simulator and from a single source of data, however they only work on supervised learning problems and require an adversary to be learned, adding computational complexity. Finally, \citet{wellmer2021ddl} concurrently explore the idea of augmenting world model dynamics for improved test-time transferability, however they focus on in-domain generalization, and do not infer context at test time.

\section{Preliminaries}
\label{sec:prelims}

We consider a Markov Decision Process (MDP), defined as a tuple $M = (\mathcal{S}, \mathcal{A}, P, R, \rho_0, \gamma)$, where $\mathcal{S}$ and $\mathcal{A}$ denote the state space and action space respectively, $P(s' | s, a)$  the transition dynamics, $R(s, a)$  the reward function, $ \rho_0$  the initial state distribution, and $\gamma \in (0, 1)$ the discount factor.
The goal in RL is to optimize a policy $\pi (a | s)$ that maximizes the expected discounted return $\mathbb{E}_{\pi, P, \rho_0}\left[\sum_{t=0}^\infty \gamma^t R(s_t, a_t)\right]$.
The value function $V^{\pi} (s) := \mathbb{E}_{\pi, P}\left[\sum_{t=0}^\infty \gamma^t R(s_t, a_t) | s_0=s\right]$ gives the expected discounted return under $\pi$ when starting from state $s$. In \textit{offline RL}, the policy is not deployed in the environment until test time. Instead, the algorithm only has access to a static dataset $\mathcal{D}_{env} = \{(s, a, r, s')\}$, collected by one or more behavioral policies $\pi_b$. We borrow notation from \cite{mopo} and refer to the distribution from which $\mathcal{D}_{env}$ was sampled as the \textit{behavioral distribution}. 

When training a model, we follow MBPO \cite{mbpo} and MOPO \cite{mopo} and train an ensemble of $N$ probabilistic dynamics models \cite{nixweigend}. Each model learns to predict both next state $s'$ and reward $r$ from a state-action pair, using $\mathcal{D}_{env}$ in a supervised fashion. Furthermore, each model outputs a Gaussian {\small $\widehat{P}_i(s_{t+1}, r_t | s_t,a_t) = \mathcal{N}(\mu(s_t, a_t), \Sigma(s_t, a_t))$}. The resulting model {\small $\widehat{P}$} defines a \textit{model MDP} {\small $\widehat{M} = (\mathcal{S}, \mathcal{A}, \widehat{P}, \widehat{R}, \rho_0, \gamma)$}, where {\small $\widehat{R}$} refers to the learned reward model. 


To train the policy, we use $k$ step rollouts inside $\widehat{M}$, adding experience to a replay buffer $\mathcal{D}_{\widehat{\text{env}}}$ to learn an action-value function and a policy, using Soft Actor Critic (SAC \cite{sac-v2}).  SAC alternates between a soft policy evaluation step, which estimates $Q^\pi(s,a)=\mathbb{E}_\pi[\sum_{t=0}^\infty\gamma^t (R(s_t, a_t) + \alpha \mathcal{H}(\pi(\cdot|s_k)))|s_0=s,a_0=a]$ using Bellman backups (where $\alpha$ is a temperature parameter for policy entropy $\mathcal{H}$), and a policy improvement which learns a policy by minimizing the expected KL divergence $J_\pi(\phi,\mathcal{D})=\mathbb{E}_{s_t\sim\mathcal{D}_{\widehat{\text{env}}}}[D_{KL}(\pi \| \mathrm{exp}\{\frac{1}{\alpha}\{Q^\pi-V^\pi\}\})]$. Note that the SAC algorithm is unchanged from the model-free setting, aside from the environment being a learned model, and the rollout horizon $k$ being truncated. Perhaps surprisingly, this approach alone produces strong results in the offline setting. However MOPO \cite{mopo} and MOReL \cite{morel} show improvement in performance by penalizing rewards in regions of the state space where the ensemble of probabilistic models is less certain. This implicitly reduces reliance on samples which deviate beyond the support of $\mathcal{D}_{env}$. 


While MOPO and MOReL have addressed the issue of training a policy in $\mathcal{D}_{env}$, and transferring to the true environment $M$, they only consider where the data in $\mathcal{D}_{env}$ is actually drawn from $P$. However, sometimes this may not be sufficient for deployment. For example, a robot could fail to walk when learning from data that was collected by a different version of the robot (with different mass), or if the same robot collected data but in a different room to deployment (with varied friction). It is this setting, where dynamics may vary at test time, that is the focus of our work. To learn successfully we propose a novel approach to training robust context-dependent policies.


\begin{figure*}[ht]
    \centering
    \includegraphics[width=0.75\textwidth]{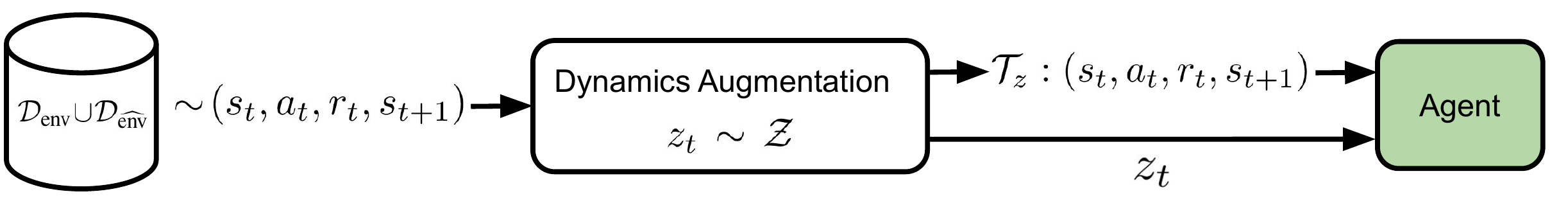}
    \vspace{-3mm}
    \caption{\small{Training policies in Augmented World Models. For each state-action pair sampled from the buffer, a new augmentation $z_t\sim\mathcal{Z}$ is sampled to produce an augmentation operator $\mathcal{T}_z$, which is applied to the transition. The policy is then trained with the new tuple of data, with the context concatenated to the state.}}
    \label{fig:train}
\end{figure*}

\begin{figure*}[ht]
    \vspace{-1mm}

    \centering
    \includegraphics[width=0.62\textwidth]{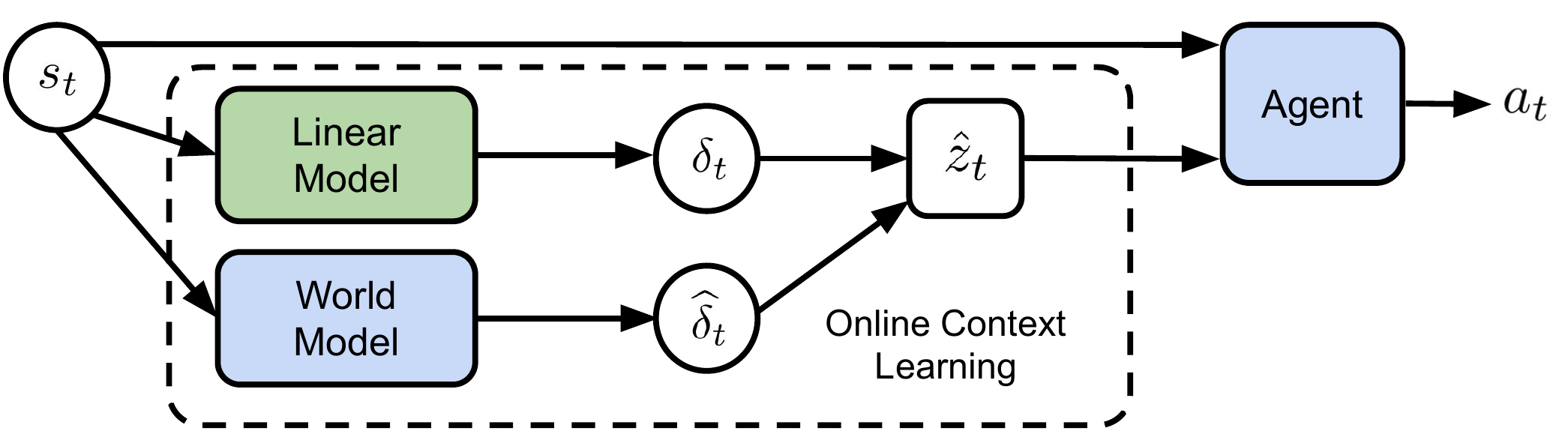}
    \vspace{-2mm}
    \caption{\small{Self-supervised policy adaptation via learned context. At each timestep, the state $s_t$ is fed into a linear model (trained online at each timestep) to predict the change in state $\delta_t$, and also passed to the fixed World Model (trained on the offline data) to predict the change in state under $\widehat{P}$, $\widehat{\delta_t}$. The approximate context is then $z_t = \sfrac{\delta_t}{\hat{\delta}_t}$, which is concatenated with $s_t$ and passed to the agent to produce an action.}}
    \label{fig:test}
    \vspace{-3mm}
\end{figure*}
\section{Augmented World Models with Self Supervised Policy Adaptation}

In this section we introduce our algorithm: Augmented World Models (AugWM). We first discuss our training procedure (Fig. \ref{fig:train}) before moving onto our self-supervised approach to online context learning (Fig. \ref{fig:test}).

\subsection{Augmented World Models}

Rather than seeking to transfer our policy from $\widehat{M}$ to $M$, we instead wish to transfer from $\widehat{M}$ to $M^\star$, where $M^\star = (\mathcal{S}, \mathcal{A}, P^\star, R^\star, \rho_0, \gamma)$ is an unseen environment with \emph{different dynamics}. Thus, our emphasis shifts to producing experience inside the world model such that our agent is able to generalize to unseen, out of distribution dynamics.  We approach this problem by training our policy in an \emph{Augmented World Model}. Formally, we denote an augmentation as $z_t \sim \mathcal{Z}, z\in\mathbb{R}^{|\mathcal{S}|}$, which is sampled at each timestep to produce an augmentation operator $\mathcal{T}_z$. $\mathcal{T}_z$ is applied to $(s,a,r,s')$ tuples from the dataset $\mathcal{D}$, and when used with tuples sampled from $\mathcal{D}_{\widehat{\mathrm{env}}}$ indirectly induces an augmented distribution {\small $\widehat{P}_z$}. In principle, we wish to produce augmentations such that the true modified environment dynamics $P^\star$ lies in the support of the distribution of augmented dynamics, i.e. 
\begin{align}
\label{eqn:requirement}
\mathrm{inf}_z D(\widehat{P}_z (s,a) \| P^\star (s,a)) \leq \epsilon
\end{align} 

for all $s, a$, some small $\epsilon>0$, and suitable distance/divergence metric $D$. We consider several augmentations, beginning with existing works before moving to new approaches which specifically target the problem of dynamics generalization. We begin with \textbf{Random Amplitude Scaling} as in \citet{rad}, which we refer to as RAD. RAD scales both $s_t$ and $s_{t+1}$ as follows:
\begin{align}
\mathcal{T}_z : (s_t, a_t, r_t, s_{t+1}) \mapsto (z \odot s_t, a_t, r_t, z\odot s_{t+1})
\end{align}
for $z \sim \mathrm{Unif}([a,b]^{|\mathcal{S}|})$. Given that our focus is on changing dynamics (vs. observational overfitting), we also propose to scale only the next state, i.e., \textbf{Random Amplitude Nextstate Scaling} (RANS): 
\begin{align}
\mathcal{T}_z: (s_t, a_t, r_t, s_{t+1}) \mapsto (s_t, a_t, r_t, z \odot s_{t+1})
\end{align}
for $z \sim \mathrm{Unif}([a,b]^{|\mathcal{S}|})$. Note that while RANS is more focused on augmenting dynamics than RAS, it still suffers from a dependence on the magnitude of $s_{t+1}$. As such, we further propose a more targeted augmentation, which we call \textbf{Dynamics Amplitude Sampling} (DAS). Rather than directly scale the state, DAS scales the \emph{change in the state} which we denote with $\delta_t = s_{t+1}-s_t$, as follows:
\begin{align}
\mathcal{T}_z : (s_t, a_t, r_t, s_{t+1}) \mapsto (s_t, a_t, r_t, s_{t} + z \odot \delta_t)
\end{align}
for $z \sim \mathrm{Unif}([a,b]^{|\mathcal{S}|})$. The full training procedure is shown in Algorithm \ref{Alg:augwm_train}.

\scalebox{0.95}{
\begin{minipage}{1\linewidth}
\vspace{-3mm}
    \begin{algorithm}[H]
    \textbf{Input:} Offline data $\mathcal{D}_{\text{env}}$, Penalty $\lambda$, horizon $h$, batchsize $B$, augmentation $\mathcal{Z}$. \\
    \textbf{Initialize:} Ensemble of $N$ dynamics models $\widehat{P}$, policy $\pi$. Replay buffer $\mathcal{D}_{\widehat{\text{env}}}$\\
    1. Train $\widehat{P}$ in a supervised fashion using $\mathcal{D}_{\text{env}}$. \\
    \For{$\text{epoch} = 1, 2, \dots$}{
       (i) Sample initial states: $\{s_1^1, \dots, s_1^B\} \sim \mathcal{D}_{\text{env}}$ \\
       (ii) Rollout policy (in parallel), using $\lambda$-penalized reward, storing all data in $\mathcal{D}_{\widehat{\text{env}}}$ \\
       (iii) Train policy using $\mathcal{D}_{\text{env}} \cup \mathcal{D}_{\widehat{\text{env}}}$. For each $(s, a, r, s')$, sample $z \sim \mathcal{Z}$ and apply $\mathcal{T}_z$.
     }

     \textbf{Return: Policy $\pi$}
     \caption{Augmented World Models: Training}
    \label{Alg:augwm_train}
    \end{algorithm}
    \vspace{0.5mm}
\end{minipage}
}

One crucial addition to our method is the use of context. Concretely, when we are optimizing the policy using a batch of data, we concatenate the next state with the augmentation vector $z$. This allows our policy to be informed of the specific augmentation that was applied to the environment and thus behave accordingly. However, at test time we do not know $z$, so what can we use? Next we propose a solution to this problem, learning the context on the fly.

\subsection{Self-Supervised Policy Selection}

In the meta-learning literature there have been many recent successes making use of a \emph{learned context} to adapt a policy at test time to a new environment \cite{pearl, varibad, zintgraf2021exploration}, typically using a blackbox model with a latent state. Crucially, these approaches require several episodes to adapt at test time, making them unfeasible in our zero-shot setting. What makes our setting unique is we explicitly know what the context represents: a linear transformation of $s_t$, or $\delta_t$. Using this insight, we are able to learn an effective context on the fly at test time. Concretely, we observe that from a state $s_t$ drawn from $M^\star$, we can sample an action $a_t\sim \pi$ and then compute an approximate $\widehat{s}_{t+1}$ using our model {\small $\widehat{P}$}. With $\widehat{s}_{t+1}$, we have a sample estimate of the \emph{state change under {\small $\widehat{P}$}}, i.e. {\small $\widehat{\delta}_t = \widehat{s}_{t+1} - s_t$}. We can make this approximation of the next state without interacting with the environment, but once we do take the action $a_t$ in the environment, we then receive the \emph{true next state} $s_{t+1}$ and can store the true difference $\delta_t = s_{t+1}-s_t$. Using the DAS augmentation, we can approximate $z$ as $\sfrac{\delta_t}{\hat{\delta}_t}$. 

\scalebox{0.95}{
\begin{minipage}{1\linewidth}
    \vspace{-2mm}
    \begin{algorithm}[H]
    \textbf{Input:} Initial state $s_1$, horizon $H$, policy $\pi$, world model $\widehat{P}$, initial context $\hat{z} = \mathbf{1}^{|\mathcal{S}|}$ \\
    \textbf{Initialize:} Linear model $f_\psi$, dataset $\mathcal{D}=\text{\O}$, return $R_0=0$. \\
    \For{$\text{step} = 1, 2, \dots \text{H}$}{
       Select action: $a_t\sim\pi(s_t, \widehat{z}_t)$\\
       Take action: $s_{t+1} \sim P^\star(s_t, a_t)$  \\
       Update return: $R_{t+1} = R_t + R^\star(s_t, a_t, s_{t+1})$ \\
       Update Dataset: $\mathcal{D} \cup (s_t, \delta_t)$. \\
       Update Linear model by minimizing $\mathcal{L}_{\text{MSE}}(\psi, \mathcal{D})$. \\
       Predict new context using $\widehat{z}_t = \sfrac{\delta_t}{{\hat{\delta}_t}}$.
     }
     \textbf{Return: $R_T$}
     \caption{Augmented World Models: Testing}
    \label{Alg:augwm_test}
    \end{algorithm}
    \vspace{0.5mm}
\end{minipage}
}

\begin{figure}[ht]
    \vspace{-2mm}
    \centering{\includegraphics[width=0.65\linewidth, trim=0 220 0 220, clip]{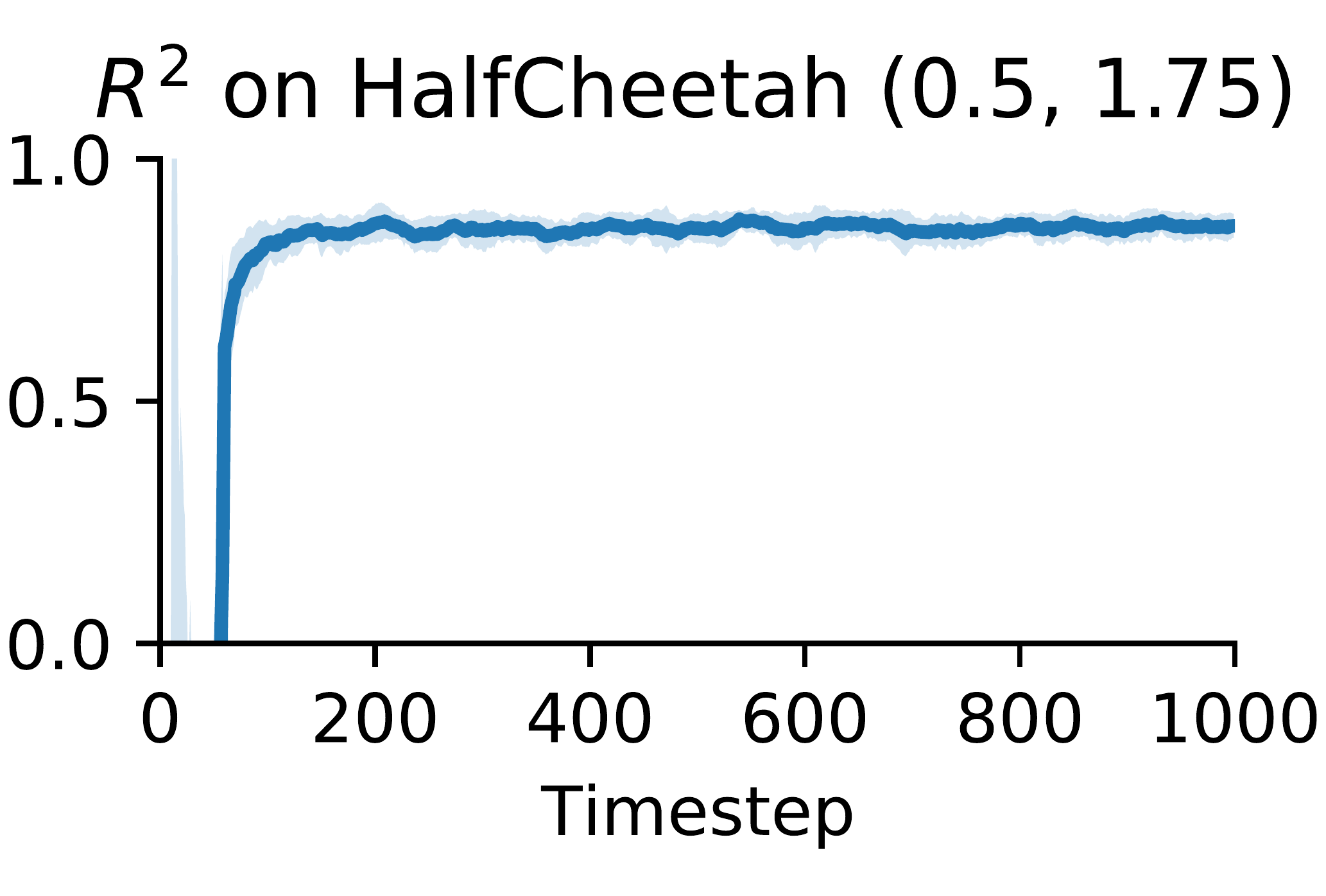}}
    \vspace{-4mm}
    \caption{\small{Mean $R^2$ of the linear model across 20 rollouts.}}
    \label{fig:r2_through_time}
    \vspace{-2mm}
\end{figure}

This however is retrospective; we can only approximate $z$ having already seen the next state, by which time our agent has already acted. Furthermore, we believe  under changed dynamics the true $z$ likely depends on $s$, thus we cannot use a previous $z$ for future timesteps. Therefore, we learn a forward dynamics model using the data collected \emph{during} the test rollout. After $h$ timesteps in the environment, we have the following dataset: $\mathcal{D} = \{(s_1, s_2), \dots, (s_{h-1}, s_h)\}$. This allows us to learn a simple dynamics model $f_\psi:(s_t) \mapsto \delta_t = s_{t+1}-s_t$, by minimizing the mean squared error $\mathcal{L}_{\text{MSE}}(\psi, \mathcal{D})$. Notably, since we never actually plan with this model, it does not need to be as accurate as a typical dynamics model in MBRL. Instead, it is crucial that the model learns quickly enough such that we can use it in a zero-shot evaluation. Thus, we choose to use a simple linear model for $f$. To show the effectiveness of this, in Fig. \ref{fig:r2_through_time} we show the mean R-squared of linear models learned on the fly during evaluation rollouts.

We observe that in less than $100$ timesteps the linear model achieves high accuracy on the test data.
Subsequently, equipped with $f_\psi$, we can approximate $\delta_t$, and predict the augmentation as $\widehat{z}_t = \sfrac{\delta_t}{{\hat{\delta}_t}}$.
We then provide the agent with $\widehat{z}_t$ to compute action $a_t$.
The full procedure is shown in Algorithm \ref{Alg:augwm_test}.


\section{Experiments}

In our experiments we aim to investigate the effectiveness of our approach for zero-shot dynamics generalization from a single offline dataset. To assess this, we will answer a series of questions, beginning with a question on the necessity of our method: 

\emph{Do we really need to develop methods specifically for dynamics generalization?}

To answer this, we train MOPO using offline data from a single environment, and test it under changed dynamics.
We consider the HalfCheetah environment from the OpenAI Gym \cite{gym}, using offline data from D4RL \cite{d4rl}. We train a MOPO agent using the mixed dataset, using our own implementation of the algorithm (but using the same hyperparameters as the original authors). To test the trained policy, we vary both the mass of the agent and damping coefficient by a multiplicative factor.\footnote{
The standard environment (both in Gym and D4RL) corresponds to both these values being set to $1.0$.}
In this work we consider a grid of the following values for HalfCheetah: $\{0.25, 0.5, 0.75, 1.0, 1.25, 1.5, 1.75\}$ and $\{0.5, 0.75, 1.0, 1.25, 1.5\}$ for Walker2d, representing a significant out-of-distribution shift. 

\begin{figure}[ht]
\centering
\vspace{-1mm}
    \centering\includegraphics[width=0.65\columnwidth, trim=0 10 0 20, clip]{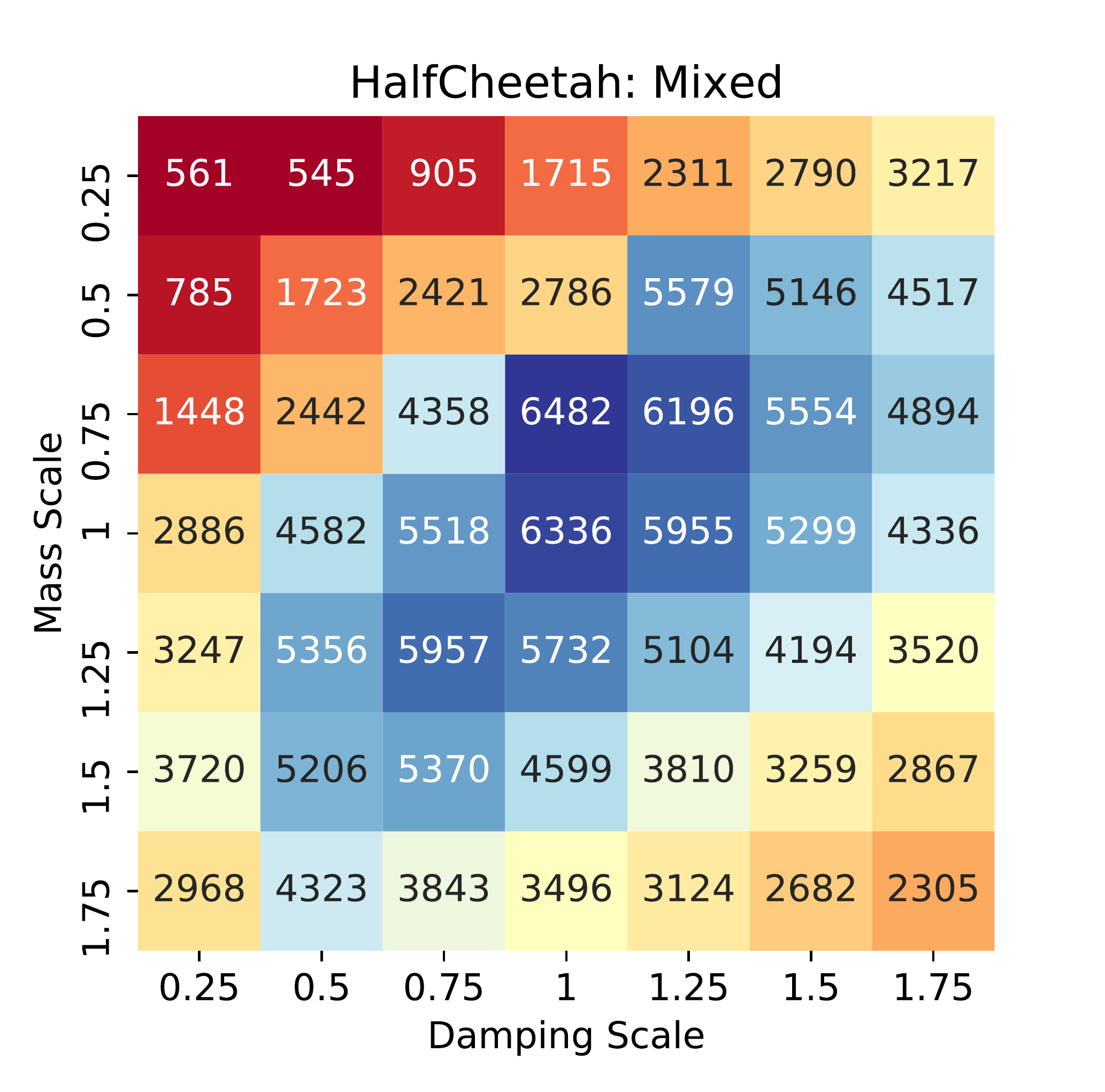}
    \vspace{-4mm}
    \caption{\small{Mean performance for 5 seeds for MOPO on HalfCheetah environments with varying dynamics. Note the central cell ($1$,$1$) corresponds to the in-sample data.}}
    \label{fig:default}
\end{figure}

The results (Fig. \ref{fig:default}), show that MOPO performance is clearly impacted by changing dynamics. We see in the central cell, that performance for our version of MOPO matches the author results \cite{mopo}, and in some cases we even see small gains (e.g. mass $=0.75$, damping $=1.0$). However, on the top left we see dramatically weaker performance, often below $1$k, indicating the robot is failing to achieve locomotion. Before evaluating AugWM, we first test whether training with the ``correct'' augmentation improves generalization performance. In short, we ask:

\emph{Is augmenting dynamics even worthwhile?}

To answer this, we train SAC for $1\times10^5$ steps and save the states visited in the `true' environment. We then use these starting states to train a policy using an offline MBPO\footnote{Since we have access to the true environment, there is no need for the MOPO penalty.} approach with AugWM. However, instead of sampling $z_t\sim\mathcal{Z}$, we provide the actual $z=\sfrac{\delta^\star}{\hat{\delta}}$ as we have access to the `true' and `modified' environments; we refer to this as an \emph{oracle} version of our method, and is designed to assess the viability of our approach. Note that we \emph{do not} augment the `true' environment rewards. We consider two baselines: a) offline MBPO in the `true' environment; b) online SAC in the `modified' environment. We train MBPO until convergence, and SAC for $1\times10^5$ steps.

\begin{figure}[H]
\centering
\vspace{-2mm}
    \centering\includegraphics[width=0.7\columnwidth]{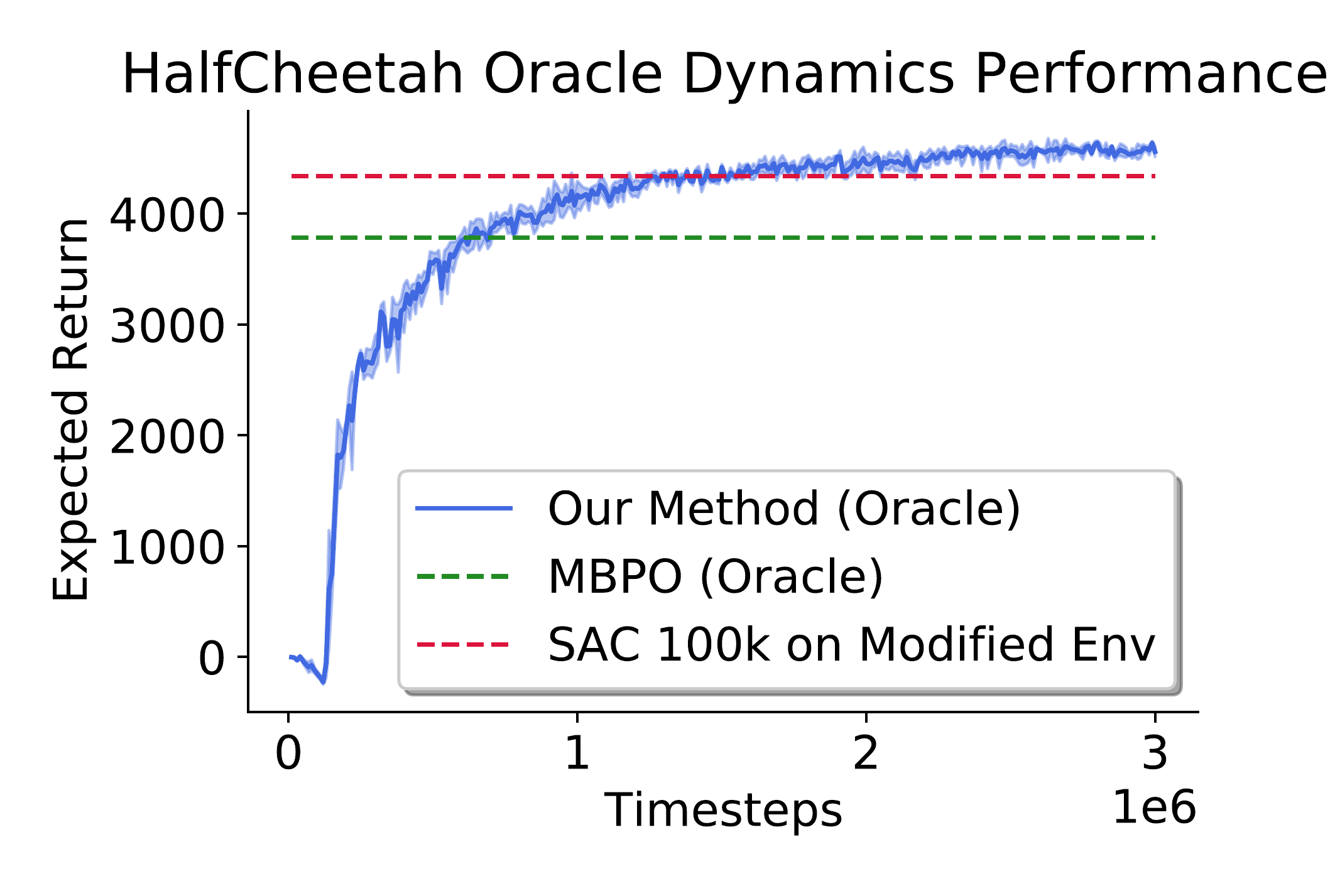}
    \vspace{-6mm}
    \caption{\small{Mean performance for 3 seeds for MBPO with a dynamics oracle on HalfCheetah with $1.5\times$ mass and damping.}}
    \label{fig:oracle}
\vspace{-4mm}
\end{figure}

As shown Fig. \ref{fig:oracle}, when provided with the true $z$, AugWM outperforms both baselines. The SAC result is surprising: the baseline agent was trained directly on the `modified' environment for the same number of steps as the policy that generated our oracle starting states. One explanation is the greater exploration induced by the `easier' dynamics of the `true' environment. This validates our approach; if we augment the dynamics {\small $\widehat{P}$} from a model correctly, we can generalize to unseen dynamics. In other words, neither the starting states nor rewards need to be from the `modified' environment. With this in mind, our next question is a simple one:

\emph{Which augmentation strategy is most effective?}

To test this, we train as in Algorithm \ref{Alg:augwm_train}, \emph{without} context, to isolate the effectiveness of the training process. We use the HalfCheetah Mixed dataset and train a MOPO agent, augmenting either both $s$ and $s'$ (RAD), just $s'$ (RANS) or just $\delta$ (DAS).

\begin{figure}[ht]
    \vspace{-1mm}
    \centering
    \includegraphics[width=0.9\columnwidth]{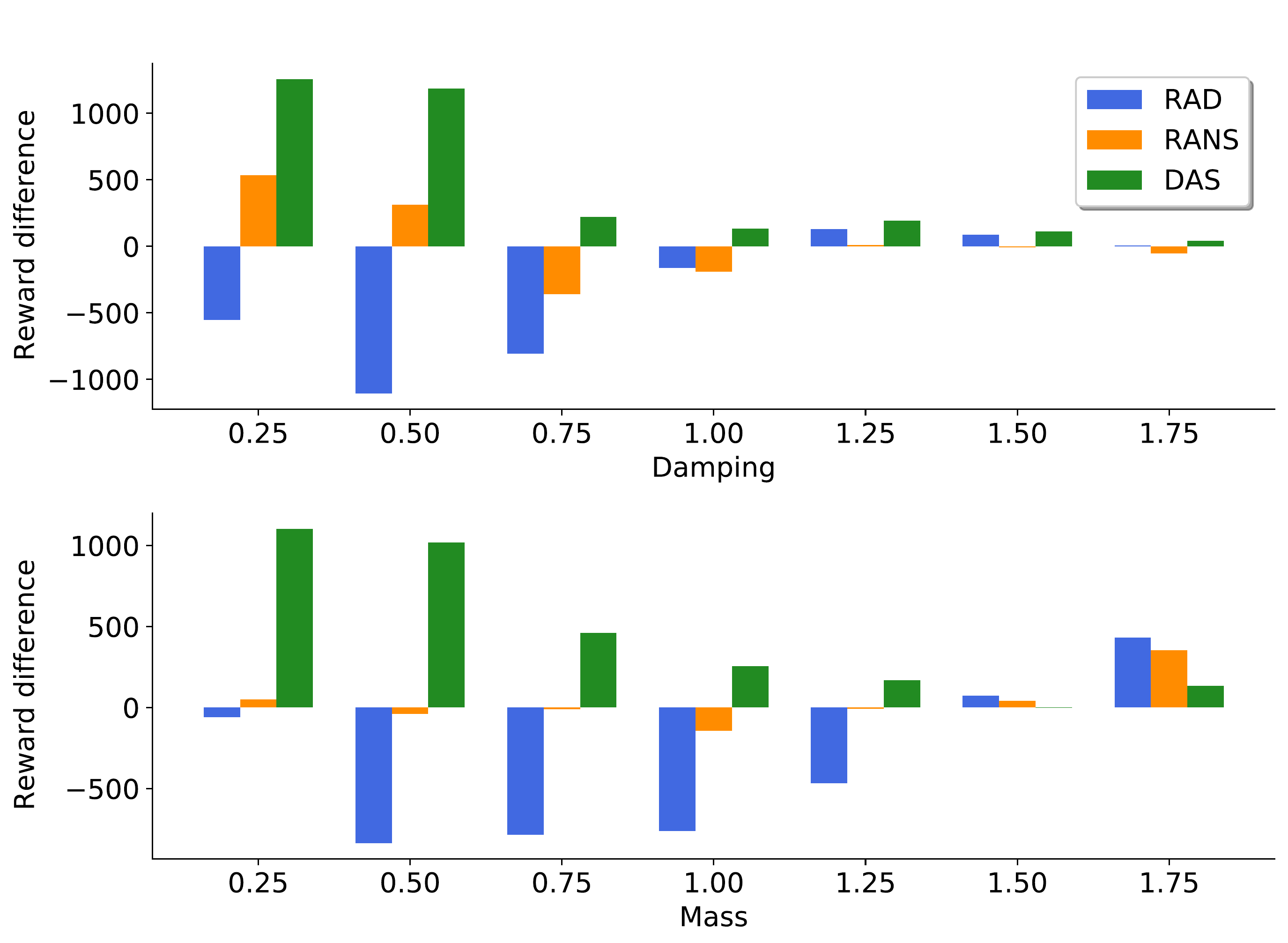}
    \vspace{-4mm}
    \caption{\small{Average performance gains of different World Model augmentations over base MOPO for different levels of damping and mass in the HalfCheetah test environment (5 seeds).}}
    \label{fig:aug_results}
    \vspace{-1mm}
\end{figure}

The results are shown in Fig. \ref{fig:aug_results}. As we see, the RAD augmentation fails to improve dynamics generalization, actually leading to worse performance overall. RANS does improve performance on unseen dynamics, as we are influencing the \emph{dynamics}, not just the observation. However, DAS clearly provides the strongest performance. As a result, we use DAS for AugWM. Our final algorithm design question is as follows: 

\emph{Does training with context improve performance?}

To answer this question, we return to the HalfCheetah Mixed setting from Fig. \ref{fig:aug_results}, taking the policy trained with DAS. We now train two additional agents: 1) Default Context: at \emph{train} time the agent is provided with the DAS augmentation $z$ as context, at \emph{test} time it is provided with a vector of ones, $\mathbf{1}^{|\mathcal{S}|}$; 2) Learned Context: trained as in 1), but context is learned online using Algorithm \ref{Alg:augwm_test}.

\begin{figure}[ht]
    \centering
    \includegraphics[width=0.9\columnwidth]{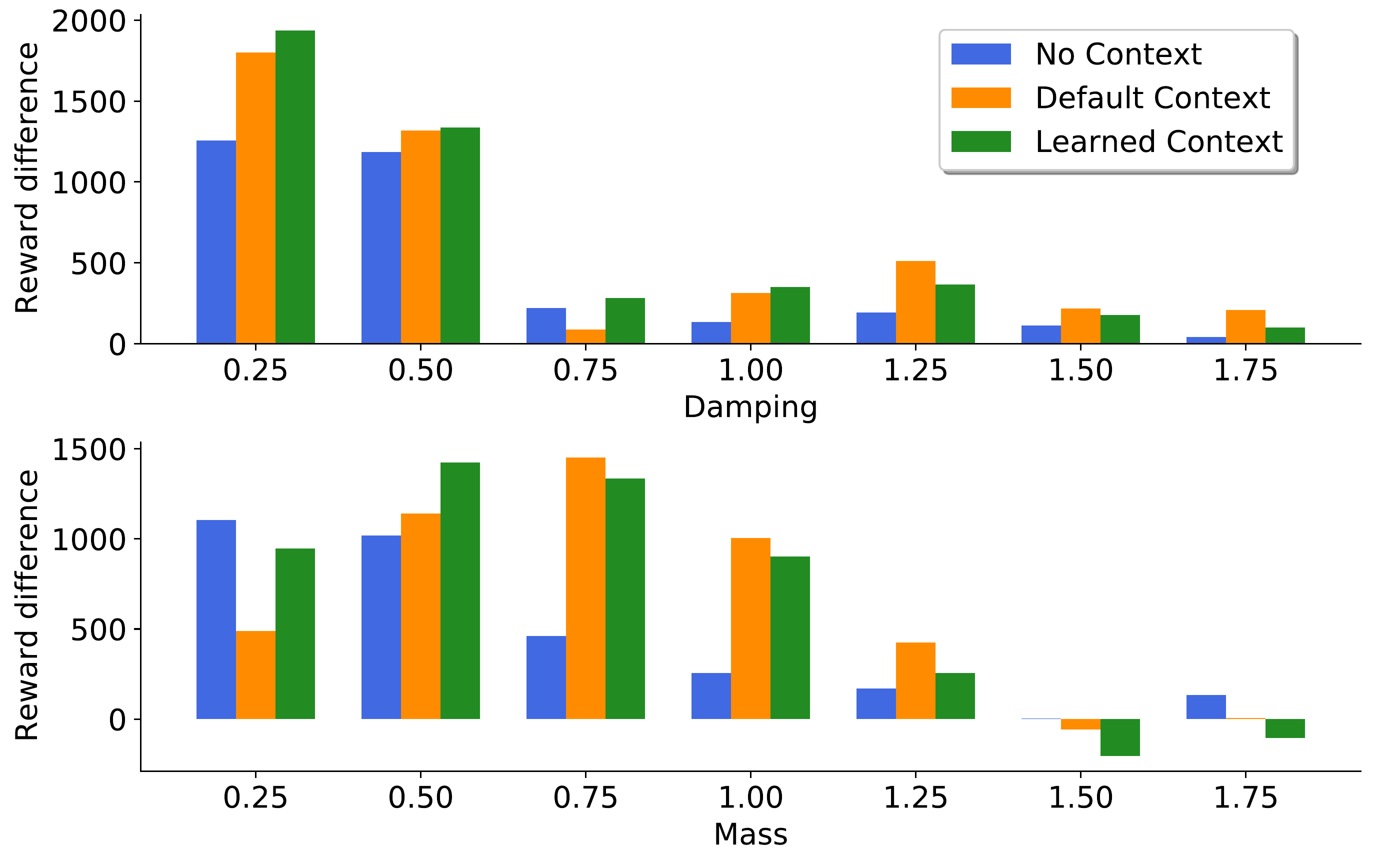}
    \vspace{-4mm}
    \caption{\small{Average performance gains of adding contexts over base MOPO for different levels of damping and mass in the HalfCheetah test environment (5 seeds).}}
    \label{fig:aug_results_context}
    \vspace{-1mm}
\end{figure}

The results are shown in in Fig. \ref{fig:aug_results_context}. We observe that training with context (orange) improves performance on average, while adapting the context on the fly (green) leads to further gains (+80 on average). These methods combine to produce our AugWM algorithm. We are now ready for the final question:

\emph{Can Augmented World Models improve zero-shot generalization?}

\begin{figure*}[h]
    \vspace{-0mm}
    \centering\includegraphics[trim=60 0 60 0, clip, width=0.9\textwidth]{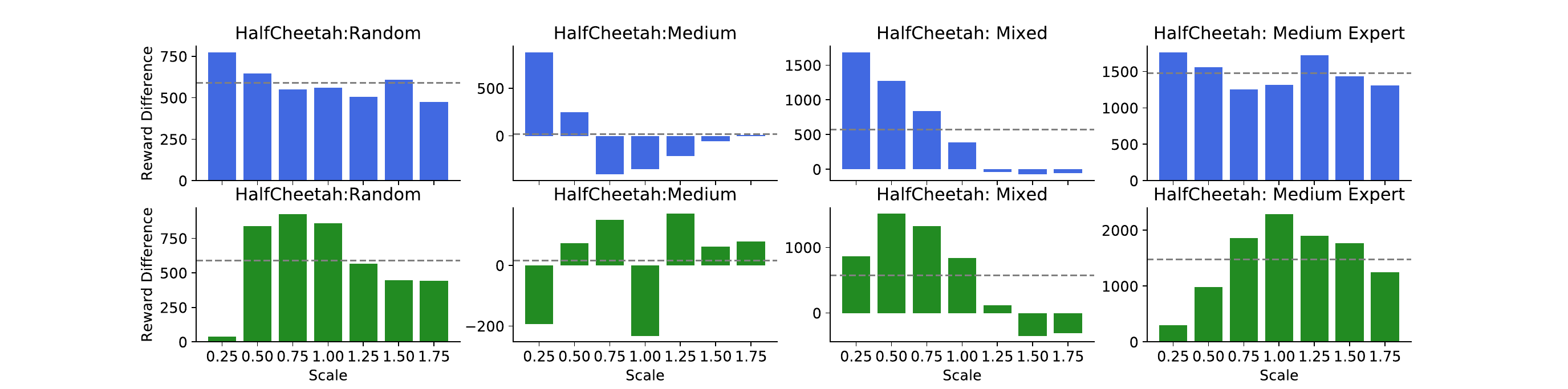}
    \vspace{-2mm}
    \caption{\small{Mean improvement for Augmented World Models over MOPO for the HalfCheetah environment, averaged over five seeds. Top row (blue) = damping scale, bottom row (green) = mass scale. Dotted line is the mean, the same value for both damping and mass.}}
    \vspace{3mm}
    \includegraphics[trim=60 0 60 0, clip, width=0.9\textwidth]{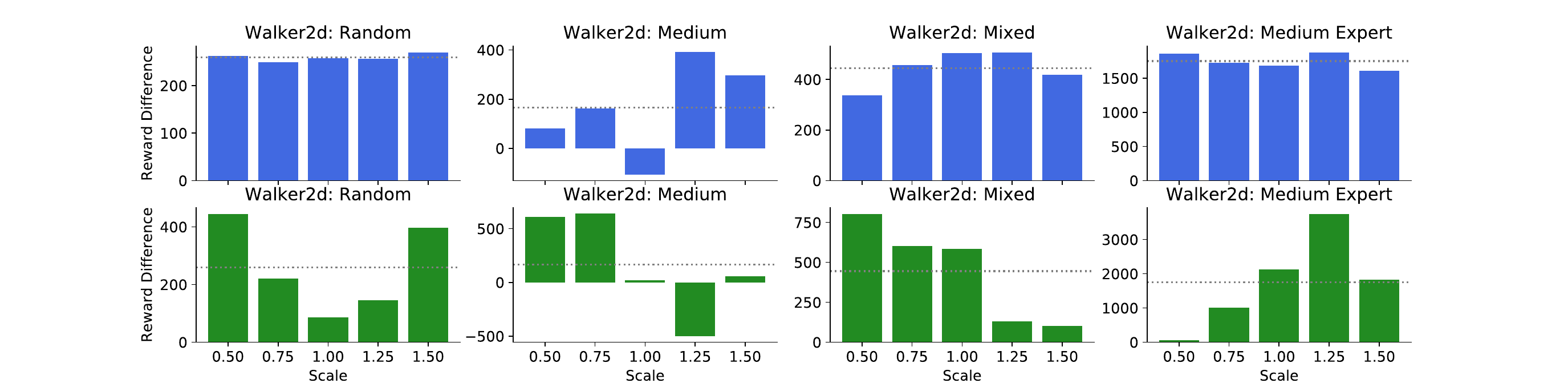}
    \vspace{-2mm}
    \caption{\small{Mean improvement for Augmented World Models over MOPO for the Walker2d environment, averaged over five seeds. Top row (blue) = damping scale, bottom row (green) = mass scale. Dotted line is the mean, the same value for both damping and mass.}}
    \label{fig:main_results}
\end{figure*}

To answer this question we perform a rigorous analysis, using multiple benchmarks from the D4RL dataset \cite{d4rl}. Namely, we consider the Random, Medium, Mixed and Medium-Expert datasets for both Walker2d and HalfCheetah. In each setting, we compare AugWM against base MOPO on zero-shot performance, training entirely on the data provided, but not seeing the test environment until evaluation. The results are shown as a change v.s.\ MOPO, averaged over one dimension in Fig. \ref{fig:main_results}, and as a total return number averaged over both dimensions in Table \ref{table:main_results}. For additional implementation details (e.g., hyperparameters) see Appendix \ref{sec:implemenation}. AugWM provides \emph{statistically significant} improvements in zero-shot performance v.s.\ MOPO in many cases, achieving successful policies where MOPO fails. 

\begin{table}[h]
\begin{minipage}[b]{0.99\linewidth}
\centering
    \vspace{-3mm}
    \caption{\small{Each entry for HalfCheetah is the mean of $49$ different dynamics, while for Walker2d it is over $25$ dynamics. Results are mean $\pm1$std. $\star$ indicates $p<0.05$ for Welch's t-test for gain over MOPO (5 seeds).}}
\scalebox{0.84}{
    \begin{tabular}{l*{4}{c}r}
    \toprule
    \textbf{Dataset Type} & \textbf{Environment} & \textbf{MOPO} & \textbf{AugWM (Ours)} \\
    \midrule
    Random & HalfCheetah & $2303 \pm112$ & $2818 \pm 197$ $\star$ \\
    Random & Walker2d & $569 \pm103$ & $706 \pm 139$\\
    \midrule
    Mixed & HalfCheetah & $3447 \pm218$ & $3948 \pm 122$ $\star$\\
    Mixed & Walker2d & $946 \pm95$ & $1317 \pm 206$ $\star$ \\
    \midrule
    Medium & HalfCheetah & $2954 \pm89$ & $2967 \pm 106$ \\
    Medium & Walker2d & $1477 \pm337$ &  $1614 \pm 440$ \\
    \midrule
    Med-Expert & HalfCheetah  & $1590 \pm766$ &  $2885 \pm 432$ $\star$ \\
    Med-Expert & Walker2d & $1062 \pm334$ & $2521 \pm 316$ $\star$ \\
    \bottomrule
    \\
\end{tabular}}%
\vspace{-3mm}
\label{table:main_results}
\end{minipage}
\vspace{-1.5mm}
\end{table}

\begin{figure*}[h!]
\centering
  ~
  \centering
  \begin{subfigure}[t]{0.19\textwidth}
    \centering\includegraphics[width=0.8\columnwidth]{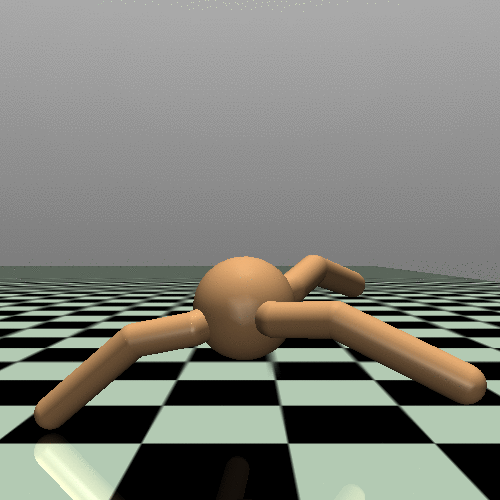}
    \caption{\small{MOPO on the regular Ant environment}}
    \label{fig:vm1}
  \end{subfigure}
  ~
  \hspace*{1cm}
  \begin{subfigure}[t]{0.19\textwidth}
    \centering\includegraphics[width=0.8\columnwidth]{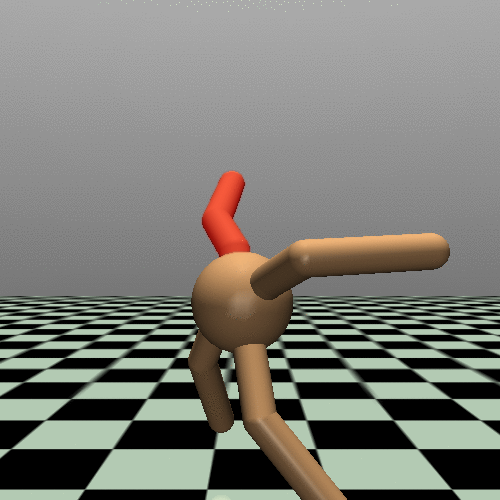}
    \caption{\small{MOPO on the crippled leg Ant environment}}
    \label{fig:vm2}
  \end{subfigure}
  ~
  \hspace*{1.0cm}
  \begin{subfigure}[t]{0.19\textwidth}
    \centering\includegraphics[width=0.8\columnwidth]{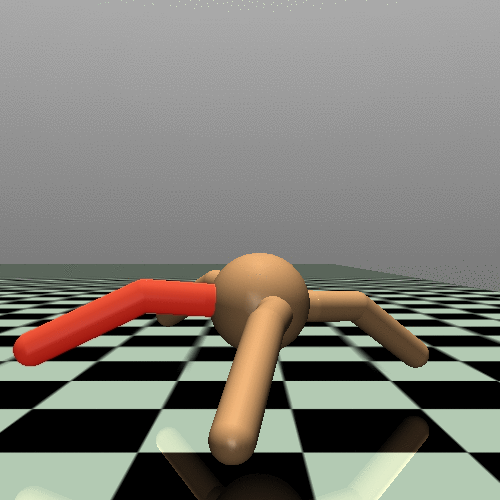}
    \caption{\small{AugWM on the crippled leg Ant environment}}
    \label{fig:vm3}
  \end{subfigure}
  \vspace{-2mm}
  \caption{\small{Comparison between MOPO and AugWM on the crippled leg Ant environment. The crippled leg is highlighted in red. The MOPO agent flips over while the AugWM agent is able to adapt to changed dynamics and successfully run. MOPO on the regular Ant environment is provided as a reference.}}
  \label{fig:visual_mjc_comparison}
  \vspace{-4mm}
\end{figure*}

By now we have provided significant evidence that AugWM can significantly improve performance for HalfCheetah and Walker2d with varied mass and damping. However, this is only a small subset of possible dynamics changes. We next consider several significantly harder settings. We test increased dimensionality, using the Ant environment from MOPO \cite{mopo}, and consider additional types of dynamics changes (e.g., Ant with crippled legs, HalfCheetah with changed limb sizes from \citet{henderson2017multitask}). We illustrate the impact of the crippled leg Ant environment on baseline agent performance, and the improvement provided by AugWM, in Fig.\ \ref{fig:visual_mjc_comparison}. We show the mean results over each of these factors of variation in Table \ref{table:mod_results}, where again AugWM provides a non-trivial improvement over a strong baseline. For more details see Appendix \ref{sec:implemenation}.

\begin{table}[h]
    \vspace{-3mm}
    \begin{minipage}{.99\linewidth}
      \caption{\small{Mean performance for MOPO, AugWM with the default context, and AugWM with learned context (LM). Entries are mean zero-shot reward for all dynamics. Bold = highest (5 seeds).}}
      \centering
        \scalebox{0.73}{
        \begin{tabular}{l*{4}{c}r}
        \toprule
         \textbf{Setting} & \textbf{MOPO}  & \textbf{AugWM (Default)} & \textbf{AugWM (LM)} \\
        \midrule
        Ant: Mass/Damp & $1634$  & $1715$ & $\mathbf{1804}$   \\
        \midrule
        Ant: One Crippled Leg & 1370   & 1572 &  \textbf{1680}   \\
        Ant: Two Crippled Legs & 700  & 697 & \textbf{795}   \\
        \midrule
        HalfCheetah: Big & 4891 & \textbf{5194} & 4968   \\
        HalfCheetah: Small  & 5151 & \textbf{5488} & 5263   \\
        \bottomrule
        \end{tabular}}
\label{table:mod_results}
    \end{minipage}%
    \vspace{-2mm}
\end{table}

\begin{figure}[H]
    \vspace{-1mm}
    \centering
    \includegraphics[width=0.85\columnwidth]{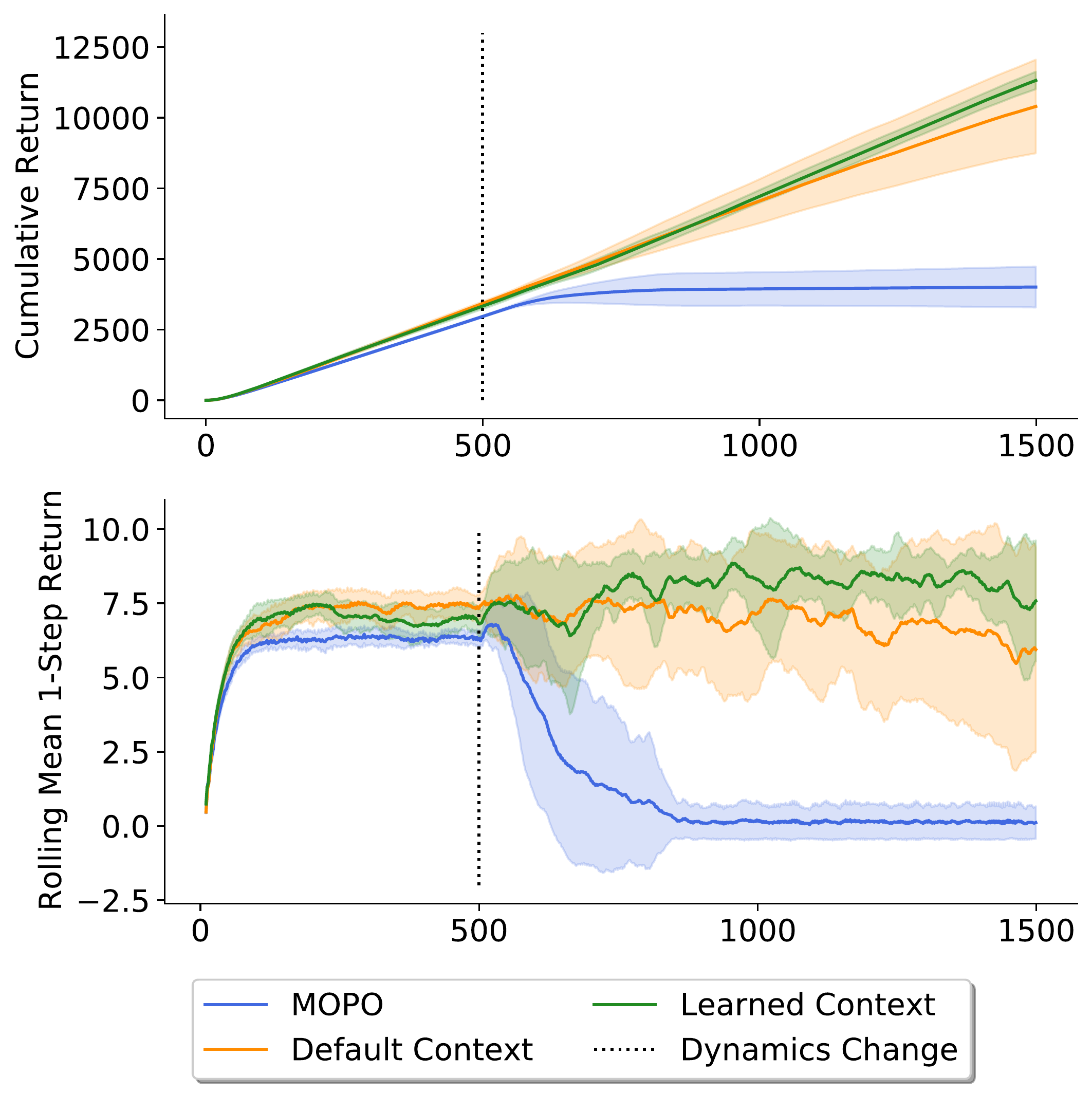}
    \vspace{-2mm}
    \caption{\small{Performance under changing dynamics. Top = cumulative returns, Bottom = rolling average single step reward. Both averaged over twenty seeds, shaded area shows $\pm1$std.}}
    \vspace{-0mm}
    \label{fig:changing_dynamics}
\end{figure}

Finally, we note that dynamics may change \emph{during} an episode; consider a robot that suffers a motor fault, reducing the power delivered to its joints. Evidently the underlying dynamics have been altered, and being robust to such changes when only training from a single dataset of offline experience is challenging. To illustrate this, we perform a $1500$ step rollout in the HalfCheetah environment, starting with offline dynamics (mass/damping = $1$), before changing to mass = $0.75$, damping = $0.5$ after $500$ steps; performance is shown in Fig. \ref{fig:changing_dynamics}. Observe that after $500$ steps, MOPO performance is dramatically reduced. This is because the agent continues to apply the same force and thus falls forward with lighter mass. For our AugWM agent, performance initially drops, then when the new context is learned we achieve \emph{higher} performance than before, making use of the lighter torso.

\textbf{Discussion} We believe that our experiments provide significant support to the claim that training with AugWM improves zero-shot dynamics generalization. In a broad set of commonly used datasets, and with a wide range of out-of-distribution dynamics, our algorithm learns good policies where a state-of-the-art baseline fails.\footnote{For videos see: \url{https://sites.google.com/view/augmentedworldmodels/}} This is due to a number of novel contributions: 1) using \emph{dynamics augmentation} rather than \emph{observation augmentation}; 2) training and testing with a context-based policy. Regarding limitations, we note that training inside the WM with context generally takes longer to converge (Appendix \ref{sec:implemenation}). Furthermore, in more nonlinear settings such as the HalfCheetah modified body part setting we saw a reduced performance for the learned context. This could be because the dynamics changes are out of the distribution of DAS augmentations (violating Eqn.\ \ref{eqn:requirement}), or due to the difficulty of modeling the task with a linear model. We note that linear models have achieved success in planning \cite{pmlr-v48-gu16} and meta learning \cite{peng2021linear}, and are effective in our case due to their data efficiency, but can be replaced by more flexible models to deal with different augmentations. Indeed, given our work is the first of its kind, we believe significant improvements are possible, such as using more complex and problem-specific augmentations.

\section{Conclusion and Future Work}

In this paper we propose Augmented World Models (AugWM), which we show sufficiently simulates changes in dynamics such that agents can generalize in a zero-shot manner. We believe that we are the first to propose this problem setting, and our results show a significant improvement over existing state-of-the-art methods which ignore this problem. 

A promising line of future work would be to meta-train a policy over AugWM such that it can quickly adapt to new dynamics in the few-shot setting. There is evidence that data augmentation can improve robustness in meta-learning \cite{rajendran2020metalearning}, and could extend to strong performance in out-of-distribution tasks. We also wish to consider varying goals at test time, and other potential sources of non-stationarity which may impact policy performance \cite{igl2021transient}. It may also be possible to extend AugWM to pixel-based tasks, which have received a great deal of recent attention \cite{planet, dreamer}. We believe that our transition based augmentations will be applicable to a latent representation, as commonly used in state-of-the-art vision MBRL approaches. Thus we think that extending our work to this setting, while a considerable feat of engineering, should not require significant methodological changes.

\section*{Acknowledgments}

Philip Ball is funded through the Willowgrove Studentship.
Cong Lu is funded by the Engineering and Physical Sciences Research Council (EPSRC). We are grateful to Taylor Killian for useful discussions on contextual/HiP-MDPs, and to Vitaly Kurin for his feedback on an earlier version of this paper via his `\href{https://www.notion.so/Augmented-World-Models-Facilitate-Zero-Shot-Dynamics-Generalization-From-a-Single-Offline-Environmen-1673b816a19b46e499ea0347c238f177}{Paper Notes}'.
The authors would also like to thank the anonymous ICLR SSL-RL Workshop + ICML reviewers and the area chair for constructive feedback which helped us in improving the paper.

\section*{Changes From ICML 2021 Proceedings}

We added additional related work that we were not originally aware of, updated the Acknowledgments section, and generally tidied up the formatting.

\bibliographystyle{apalike}
\bibliography{refs}

\newpage
\appendix
\onecolumn
\section*{Appendix}
\section{Additional Experiments}
\label{sec:ablation}



\begin{figure}[ht]
    \centering\includegraphics[width=.7\linewidth]{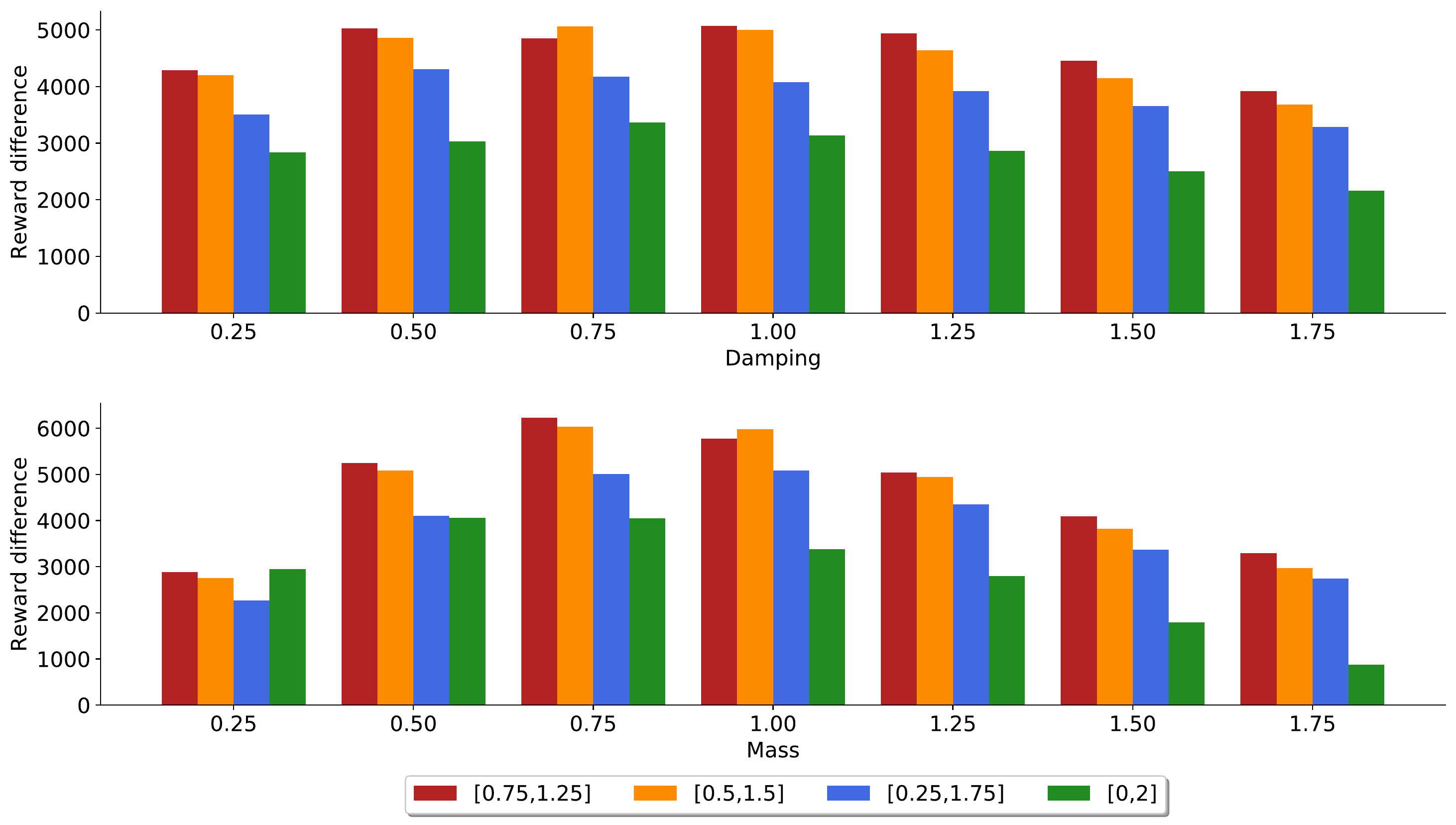}
    \vspace{-3mm}
    \caption{\small{Performance for Augmented World Models with the DAS augmentation. Each plot shows different values for $a$ and $b$, the ranges for the sampled noise.}}
    \label{fig:ablation_das}
\end{figure}

In this section we show the performance for Augmented World Models with different training ranges for the DAS augmentation ($z$ train in Table \ref{tab:awm-hyperparams}). We train with adaptive context on the HalfCheetah mixed dataset, and present the results in Fig.\ \ref{fig:ablation_das}. As we see, $[0.75,1.25]$ and $[0.5,1.5]$ perform the best. Based on this, we use $[0.5,1.5]$ for our experiments as we believe this helps us sample a wider set of dynamics, helping us generalize better across all environments and data sets.

\section{Implementation Details}
\label{sec:implemenation}

\subsection{Hyperparameters}

Our algorithm is based on MOPO \cite{mopo} with values for the rollout length $h$ and penalty coefficient $\lambda$ shown in Table \ref{tab:mopo-hyperparam}.
\begin{table}[ht]
\centering
\caption{Hyperparameters used in the D4RL datasets.}
\label{tab:mopo-hyperparam}
\begin{tabular}{@{}llc@{}}
\toprule
\textbf{Dataset Type} & \textbf{Environment} & \multicolumn{1}{l}{\textbf{MOPO ($h$, $\lambda$)}} \\ \midrule
random     & halfcheetah & 5, 0.5 \\
random     & walker2d    & 1, 1   \\
medium     & halfcheetah & 1, 1   \\
medium     & walker2d    & \hspace{0.14cm} 1, 1 \tablefootnote{We follow the original MOPO hyperparameters for all datasets except for walker2d-medium where we found (1, 1) worked better for both MOPO and our method than (5, 5).}   \\
mixed      & halfcheetah & 5, 1   \\
mixed      & walker2d    & 1, 1  \\
med-expert & halfcheetah & 5, 1   \\
med-expert & walker2d    & 1, 2   \\ \bottomrule
\end{tabular}
\end{table}

AugWM specific hyperparameters are listed in Table \ref{tab:awm-hyperparams}. For each evaluation rollout, we clear the buffer of stored true modified environment transitions to measure zero-shot performance. We adapt using the context after a set number of steps, $k$, in the environment to train the linear model. The two ranges used for the context $z$ during training and test time are different. At test time, the estimated context is clipped to remain within the given bounds.
\begin{table}[H]
\centering
\caption{AugWM Hyperparameters}
\label{tab:awm-hyperparams}
\begin{tabular}{@{}lc@{}}
\toprule
\textbf{Parameter}           & \textbf{Value}   \\ \midrule
evaluation rollouts          & 5                \\
MOPO offline epochs          & 400              \\
AugWM offline epochs           & 900              \\
$k$, steps for adaptation    & 300              \\
$z$ train range              & {[}0.5, 1.5{]}   \\
$z$ test range               & {[}0.93, 1.07{]} \\ \bottomrule
\end{tabular}
\end{table}

\subsection{D4RL dataset}
We evaluate our method on D4RL \cite{d4rl} datasets based on the MuJoCo continuous control tasks (halfcheetah and walker2d). The four dataset types we evaluate on are:
\begin{itemize}
    \item \textbf{random:} roll out a randomly initialized policy for 1M steps.
    \item \textbf{medium:} partially train a policy using SAC, then roll it out for 1M steps.
    \item \textbf{mixed:} train a policy using SAC until a certain (environment-specific) performance threshold is reached, and take the replay buffer as the batch.
    \item \textbf{medium-expert:}  combine 1M samples of rollouts from a fully-trained policy with another 1M samples of rollouts from a partially trained policy or a random policy.
\end{itemize}
This gives us a total of 8 experiments.

\subsection{Ant Environment}

For the Ant experiments, we follow the Ant Changed Direction approach in MOPO \cite{mopo}. Since this offline dataset is not provided in the authors' code, nor is it in the standard D4RL library \cite{d4rl}, we were required to generate our own offline Ant dataset. Since the authors' did not outline certain details in their experiment, we found the following was required to match their performance with our codebase: 1) Training our SAC policy for $1\times10^6$ timesteps in the Ant environment provided by the authors' code in \cite{mopo}; 2) relabelling each reward in the buffer using the new direction, without the living reward; 3) training a world model over this offline dataset; 4) training a policy in the world model, adding in living reward post-hoc; 5) evaluating the policy with the living reward.

\subsection{HalfCheetah Modified Agent}

We use the modified HalfCheetah environments from \cite{henderson2017multitask}. In each setting one body part of the agent is changed, from following set: \{Foot, Leg, Thigh, Torso, Head\}. The body part can either be ``Big'' or ``Small'', where Big bodyparts involve scaling the mass and width of the limb by 1.25 and Small bodyparts are scaled by 0.75. In Table \ref{table:mod_results} we show the mean over each of these five body parts, for agents trained on each of the D4RL datasets, repeated for five seeds.

\end{document}